\documentclass[journal]{IEEEtran}

\usepackage{cite}
\usepackage[pdftex]{graphicx}
\usepackage{adjustbox}
\usepackage{amsmath}
\usepackage{amssymb}
\usepackage{array}
\usepackage{multirow}
\usepackage{color, soul}
\soulregister\ref7

\newcommand*\rot{\rotatebox{90}}

\begin{document}

\title{Semantic Structure and Interpretability of \\ Word Embeddings}

\author{L\"{u}tfi~Kerem~\c{S}enel,~\IEEEmembership{Student Member,~IEEE,}
        \.{I}hsan~Utlu,
        Veysel Y\"{u}cesoy,~\IEEEmembership{Student Member,~IEEE,}
        \\Aykut Ko\c{c},~\IEEEmembership{Member,~IEEE,}
        and Tolga \c{C}ukur,~\IEEEmembership{Senior Member,~IEEE,}

\thanks{L. K. \c{S}enel, \.{I}. Utlu, V. Y\"{u}cesoy and A. Ko\c{c} are with ASELSAN Research Center, Ankara, Turkey (e-mail: lksenel@aselsan.com.tr).
	
	L. K. \c{S}enel, \.{I}. Utlu, V. Y\"{u}cesoy and T. \c{C}ukur are with Electrical and Electronics Engineering, Bilkent University, Ankara, Turkey.
	
	L. K. \c{S}enel and T. \c{C}ukur are with UMRAM, Bilkent  University, Ankara, Turkey.
	
	T. \c{C}ukur is with Neuroscience Program, Sabuncu Brain Research Center, Bilkent  University, Ankara, Turkey.}

\thanks{T. \c{C}ukur and A. Ko\c{c} mutually supervised this work under a joint industry-university co-advising program.}

\thanks{\copyright 2018 IEEE. Personal use of this material is permitted. Permission from IEEE must be obtained for all other uses, including reprinting/republishing this material for advertising or promotional purposes, collecting new collected works for release or redisibution to servers or lists, or reuse of any copyrighted component of this work in other works.}}

%

\maketitle

\begin{abstract}
	
Dense word embeddings, which encode meanings of words to low dimensional vector spaces, have become very popular in natural language processing (NLP) research due to their state-of-the-art performances in many NLP tasks. Word embeddings are substantially successful in capturing semantic relations among words, so a meaningful semantic structure must be present in the respective vector spaces. However, in many cases, this semantic structure is broadly and heterogeneously distributed across the embedding dimensions making interpretation of dimensions a big challenge. In this study, we propose a statistical method to uncover the underlying latent semantic structure in the dense word embeddings. To perform our analysis, we introduce a new dataset (SEMCAT) that contains more than 6,500 words semantically grouped under 110 categories. We further propose a method to quantify the interpretability of the word embeddings. The proposed method is a practical alternative to the classical word intrusion test that requires human intervention.

\end{abstract}


\IEEEpeerreviewmaketitle

\section{Introduction}

\IEEEPARstart{W}{ords} are the smallest elements of a language with a practical meaning. Researchers from diverse fields including linguistics \cite{miller1995wordnet}, computer science \cite{mikolov2013b} and statistics \cite{bordes2012} have developed models that seek to capture  ``word meaning" so that these models can accomplish various NLP tasks such as parsing, word sense disambiguation and machine translation. Most of the effort in this field is based on the distributional hypothesis \cite{harris1954} which claims that a word is characterized by the company it keeps \cite{firth1957}. Building on this idea, several vector space models such as well known Latent Semantic Analysis (LSA) \cite{deerwester1990lsa} and Latent Dirichlet Allocation (LDA) \cite{blei2003lda} that make use of word distribution statistics have been proposed in distributional semantics. Although these methods have been commonly used in NLP, more recent techniques that generate dense, continuous valued vectors, called \textit{embeddings}, have been receiving increasing interest in NLP research. Approaches that learn embeddings include neural network based predictive methods \cite{mikolov2013b, bojanowski2016fasttext} and count-based matrix-factorization methods \cite{pennington2014}. Word embeddings brought about significant performance improvements in many intrinsic NLP tasks such as analogy or semantic textual similarity tasks, as well as downstream NLP tasks such as part-of-speech (POS) tagging \cite{lin2015POS}, named entity recognition \cite{sienvcnik2015NER}, word sense disambiguation \cite{iacobacci2016Disambiguation}, sentiment analysis \cite{yu2017Sentiment} and cross-lingual studies \cite{senel2017crossLingual}.

Although high levels of success have been reported in many NLP tasks using word embeddings, the individual embedding dimensions are commonly considered to be uninterpretable \cite{levy2014dependency}. Contrary to some earlier sparse vector space models such as Hyperspace Analogue to Language (HAL) \cite{lund1996HAL}, what is represented in each dimension of word embeddings is often unclear, rendering them a black-box approach. In contrast, embedding models that yield dimensions that are more easily interpretable in terms of the captured information can be better suited for NLP tasks that require semantic interpretation, including named entity recognition and retrieval of semantically related words. Model interpretability is also becoming increasingly relevant from a regulatory standpoint, as evidenced by the recent EU regulation that grants people with a ``right to explanation" regarding automatic decision making algorithms \cite{goodman2016European}. 

Although word embeddings are a dominant part of NLP research, most studies aim to maximize the task performance on standard benchmark tests such as MEN \cite{bruni2014multimodal} or Simlex-999 \cite{hill2016simlex}. While improved test performance is undoubtedly beneficial, an embedding with enhanced performance does not necessarily reveal any insight about the semantic structure that it captures. A systematic assessment of the semantic structure intrinsic to word embeddings would enable an improved understanding of this popular approach, would allow for comparisons among different embeddings in terms of interpretability and potentially motivate new research directions. 

In this study, we aim to bring light to the semantic concepts implicitly represented by various dimensions of a word embedding. To explore these hidden semantic structures, we leverage the category theory \cite{murphy2004} that defines a category as a grouping of concepts with similar properties. We use human-designed category labels to ensure that our results and interpretations closely reflect human judgements. Human interpretation can make use of any kind of semantic relation among words to form a semantic group (category). This does not only significantly increase the number of possible categories but also makes it difficult and subjective to define a category. Although several lexical databases such as WordNet \cite{miller1995wordnet} have a representation for relations among words, they do not provide categories as needed for this study. Since there is no gold standard for semantic word categories to the best of our knowledge, we introduce a new category dataset where more than 6,500 different words are grouped in 110 semantic categories. Then, we propose a method based on distribution statistics of category words within the embedding space in order to uncover the semantic structure of the dense word vectors. We apply quantitative and qualitative tests to substantiate our method. Finally, we claim that the semantic decomposition of the embedding space can be used to quantify the interpretability of the word embeddings without requiring any human effort unlike the word intrusion test \cite{chang2009}. 

This paper is organized as follows: Following a discussion of related work in Section \ref{related_word}, we describe our methods in Section \ref{methods}. In this section we introduce our dataset and also describe methods we used to investigate the semantic decomposition of the embeddings, to validate our findings and to measure the interpretability. In Section \ref{results}, we present the results of our experiments and finally  we conclude the paper in Section \ref{conclusion}.

\section{Related Work}\label{related_word}


In the word embedding literature, the problem of interpretability has been approached via several different routes. For learning sparse, interpretable word representations from co-occurrence variant matrices, \cite{murphy2012} suggested algorithms based on non-negative matrix factorization (NMF) and the resulting representations are called non-negative sparse embeddings (NNSE). To address memory and scale issues of the algorithms in \cite{murphy2012}, \cite{ luo2015} proposed an online method of learning interpretable word embeddings. In both studies, interpretability was evaluated using a word intrusion test introduced in \cite{chang2009}. The word intrusion test is expensive to apply since it requires manual evaluations by human observers separately for each embedding dimension. As an alternative method to incorporate human judgement, \cite{fyshe2014} proposed joint non-negative sparse embedding (JNNSE), where the aim is to combine text-based similarity information among words with brain activity based similarity information to improve interpretability. Yet, this approach still requires labor-intensive collection of neuroimaging data from multiple subjects. 

Instead of learning interpretable word representations directly from co-occurrence matrices, \cite{arora2016} and \cite{faruqui2015} proposed to use sparse coding techniques on conventional dense word embeddings to obtain sparse, higher dimensional and more interpretable vector spaces. However, since the projection vectors that are used for the transformation are learned from the word embeddings in an unsupervised manner, they do not have labels describing the corresponding semantic categories. Moreover, these studies did not attempt to enlighten the dense word embedding dimensions, rather they learned new high dimensional sparse vectors that perform well on specific tests such as word similarity and polysemy detection. In \cite{faruqui2015}, interpretability of the obtained vector space was evaluated using the word intrusion test. An alternative approach was proposed in \cite{zobnin2017rotations}, where interpretability was quantified by the degree of clustering around embedding dimensions and orthogonal transformations were examined to increase interpretability while preserving the performance of the embedding. Note, however, that it was shown in \cite{zobnin2017rotations} that total interpretability of an embedding is constant under any orthogonal transformation and it can only be redistributed across the dimensions. With a similar motivation to \cite{zobnin2017rotations}, \cite{park2017rotated} proposed rotation algorithms based on exploratory factor analysis (EFA) to preserve the expressive performance of the original word embeddings while improving their interpretability. In \cite{park2017rotated}, interpretability was calculated using a distance ratio (DR) metric that is effectively proportional to the metric used in \cite{zobnin2017rotations}. Although interpretability evaluations used in \cite{zobnin2017rotations} and \cite{park2017rotated} are free of human effort, they do not necessarily reflect human interpretations since they are directly calculated from the embeddings.

Taking a different perspective, a recent study, \cite{jang2017}, attempted to elucidate the semantic structure within NNSE space by using categorized words from the HyperLex dataset \cite{vulic2016hyperlex}. The interpretability levels of embedding dimensions were quantified based on the average values of word vectors within categories. However, HyperLex is constructed based on a single type of semantic relation (hypernym) and average number of words representing a category is significantly low ($\approx2$) making it challenging to conduct a comprehensive analysis.

\section{Methods}\label{methods}

To address the limitations of the approaches discussed in Section \ref{related_word}, in this study we introduce a new conceptual category dataset. Based on this dataset, we propose statistical methods to capture the hidden semantic concepts in word embeddings and to measure the interpretability of the embeddings.

\subsection{Dataset}\label{dataset}

Understanding the hidden semantic structure in dense word embeddings and providing insights on interpretation of their dimensions are the main objectives of this study. Since embeddings are formed via unsupervised learning on unannotated large corpora, some conceptual relationships that humans anticipate may be missed and some that humans do not anticipate may be formed in the embedding space \cite{gladkova2016intrinsic}. Thus, not all clusters obtained from a word embedding space will be interpretable. Therefore, using the clusters in the dense embedding space might not take us far towards interpretation. This observation is also rooted in the need for human judgement in evaluating interpretability.   

To provide meaningful interpretations for embedding dimensions, we refer to the category theory \cite{murphy2004} where concepts with similar semantic properties are grouped under a common category. As mentioned earlier, using clusters from the embedding space as categories may not reflect human expectations accurately, hence having a basis based on human judgements is essential for evaluating interpretability. In that sense, semantic categories as dictated by humans can be considered a gold standard for categorization tasks since they directly reflect human expectations. Therefore, using supervised categories can enable a proper investigation of the word embedding dimensions. In addition, by comparing the human-categorized semantic concepts with the unsupervised word embeddings, one can acquire an understanding of what kind of concepts can or cannot be captured by the current state-of-the-art embedding algorithms.

In the literature, the concept of category is commonly used to indicate super-subordinate (hyperonym-hyponym) relations where words within a category are types or examples of that category. For instance, the furniture category includes words for furniture names such as bed or table. The HyperLex category dataset \cite{vulic2016hyperlex}, which was used in \cite{jang2017} to investigate embedding dimensions, is constructed based on this type of relation that is also the most frequently encoded relation among sets of synonymous words in the WordNet database \cite{miller1995wordnet}. However, there are many other types of semantic relations such as meronymy (part-whole relations), antonymy (opposite meaning words), synonymy (words having the same sense) and cross-Part of Speech (POS) relations (i.e. lexical entailments). Although WordNet provides representations for a subset of these relations, there is no clear procedure for constructing unified categories based on multiple different types of relations. It remains unclear what should be considered as a category, how many categories there should be, how narrow or broad they should be, and which words they should contain. Furthermore, humans can group words by inference, based on various physical or numerical properties such as color, shape, material, size or speed, increasing the number of possible groups almost unboundedly. For instance, words that may not be related according to classical hypernym or synonym relations might still be grouped under a category due to shared physical properties: sun, lemon and honey are similar in terms of color; spaghetti, limousine and sky-scanner are considered as tall; snail, tractor and tortoise are slow. 

\begin{table}
	\begin{center}
		\caption{\label{tab:SEMCAT statistics} Summary Statistics of SEMCAT and HyperLex}
		\adjustbox{max width=\columnwidth}{
			\renewcommand{\arraystretch}{1.3}
			\begin{tabular}{|c|c|c|}	
				\hline  
				& SEMCAT	& HperLex  \\ \hline
				Number of Categories            	& 110   	& 1399     \\ \hline
				Number of Unique Words              & 6559      & 1752 	   \\ \hline
				Average Word Count per Category	    & 91        & 2        \\ \hline
				Standard Deviation of Word Counts   & 56        & 3         \\ \hline
				
			\end{tabular}
		}
	\end{center}
\end{table}

\begin{table}
	\begin{center}
		\caption{\label{tab:SemCat} 10 sample words from each of the 6 representative SEMCAT categories. }
		\adjustbox{max width=\columnwidth}{
			\renewcommand{\arraystretch}{1.3}
			\begin{tabular}{|c|c|c|c|c|c|}
				\hline							        	
				\bf Science		& \bf Sciences		& \bf Art 		& \bf Car 		& \bf Cooking 	& \bf Geography   \\ \hline
				atom	 		& astronomy			& abstract		& auto 			& bake			& africa 		  \\ \hline
				cell			& botany			& artist		& car			& barbeque		& border		  \\ \hline
				chemical		& economics			& brush			& convertible	& boil			& capital		  \\ \hline
				data			& genetics			& composition	& hybrid		& dough			& cartography	  \\ \hline
				element			& linguistics 		& draw			& jeep			& grill			& continent		  \\ \hline
				evolution		& neuroscience		& masterpiece	& limo			& juice			& earth			  \\ \hline
				laboratory		& psychology		& photograph 	& runabout		& marinate		& east			  \\ \hline
				microscope		& taxonomy			& perspective	& rv			& oil			& gps			  \\ \hline
				scientist		& thermodynamics	& sketch		& taxi			& roast			& river			  \\ \hline
				theory			& zoology			& style			& van			& serve			& sea			  \\ \hline
				
			\end{tabular}
		}
	\end{center}
\end{table}

In sum, diverse types of semantic relationships or properties can be leveraged by humans for semantic interpretation. Therefore, to investigate the semantic structure of the word embedding space using categorized words, we need categories that represent a broad variety of distinct concepts and distinct types of relations. To the best of our knowledge, there is no comprehensive word category dataset that captures the many diverse types of relations mentioned above. What we have found closest to the required dataset are the online categorized word-lists\footnote{www.enchantedlearning.com/wordlist/} that were constructed for educational purposes. There are a total of 168 categories on these word-lists. To build a word-category dataset suited for assessing the semantic structure in word embeddings, we took these word-lists as a foundational basis. We filtered out words that are not semantically related but share a common nuisance property such as their POS tagging (verbs, adverbs, adjectives etc.) or being compound words. Several categories containing proper words or word phrases such as the chinese new year and good luck symbols categories, which we consider too specific, are also removed from the dataset. Vocabulary is limited to the most frequent 50,000 words, where frequencies are calculated from English Wikipedia, and words that are not contained in this vocabulary are removed from the dataset. We call the resulting semantically grouped word dataset ``SEMCAT\footnote{github.com/avaapm/SEMCATdataset2018}" (SEMantic CATegories). Summary statistics of SEMCAT and HyperLex datasets are given in Table \ref{tab:SEMCAT statistics}. 10 sample words from each of 6 representative SEMCAT categories are given in Table \ref{tab:SemCat}. 

\subsection{Semantic Decomposition} \label{m:semantic_decomposition}

In this study, we use GloVe \cite{pennington2014} as the source algorithm for learning dense word vectors. The entire content of English Wikipedia is utilized as the corpus. In the preprocessing step, all non-alphabetic characters (punctuations, digits, etc.) are removed from the corpus and all letters are converted to lowercase. Letters coming after apostrophes are taken as separate words (\verb|she'll| becomes \verb|she ll|). The resulting corpus is input to the GloVe algorithm. Window size is set to 15, vector length is chosen to be 300 and minimum occurrence count is set to 20 for the words in the corpus. Default values are used for the remaining parameters. The word embedding matrix, $\mathcal{E}$, is obtained from GloVe after limiting vocabulary to the most frequent 50,000 words in the corpus (i.e. $\mathcal{E}$ is 50,000$\times$300). The GloVe algorithm is again used for the second time on the same corpus generating a second embedding space, $\mathcal{E}^2$, to examine the effects of different initializations of the word vectors prior to training. 


To quantify the significance of word embedding dimensions for a given semantic category, one should first understand how a semantic concept can be captured by a dimension, and then find a suitable metric to measure it. \cite{jang2017} assumed that a dimension represents a semantic category if the average value of the category words for that dimension is above an empirical threshold, and therefore took that average value as the representational power of the dimension for the category. Although this approach may be convenient for NNSE, directly using the average values of category words is not suitable for well-known dense word embeddings due to several reasons. First, in dense embeddings it is possible to encode in both positive and negative directions of the dimensions making a single threshold insufficient. In addition, different embedding dimensions may have different statistical characteristics. For instance, average value of the words from the jobs category of SEMCAT is around 0.38 and 0.44 in $221^{st}$ and $57^{th}$ dimensions of $\mathcal{E}$ respectively; and the average values across all vocabulary are around 0.37 and -0.05 respectively for the two dimensions. Therefore, the average value of 0.38 for the jobs category may not represent any encoding in the $221^{st}$ dimension since it is very close to the average of any random set of words in that dimension. In contrast, an average of similar value 0.44 for the jobs category may be highly significant for the $57^{th}$ dimension. Note that focusing solely on average values might be insufficient to measure the encoding strength of a dimension for a semantic category. For instance, words from the car category have an average  of -0.08 that is close to the average across all vocabulary, -0.04, for the $133^{th}$ embedding dimension. However, standard deviation of the words within the car category is 0.15 which is significantly lower than the standard deviation of all vocabulary, 0.35, for this particular dimension. In other words, although average of words from the car category is very close to the overall mean, category words are more tightly grouped compared to other vocabulary words in the $133^{th}$ embedding dimension, potentially implying significant encoding.

From a statistical perspective, the question of ``How strong a particular concept is encoded in an embedding dimension?" can be interpreted as ``How much information can be extracted from a word embedding dimension regarding a particular concept?". If the words representing a concept (i.e. words in a SEMCAT category) are sampled from the same distribution with all vocabulary words, then the answer would be zero since the category would be statistically equivalent to a random selection of words. For dimension $i$ and category $j$, if $\mathcal{P}_{i,j}$ denotes the distribution from which words of that category are sampled and $\mathcal{Q}_{i,j}$ denotes the distribution from which all other vocabulary words are sampled, then the distance between distributions $\mathcal{P}_{i,j}$ and $\mathcal{Q}_{i,j}$ will be proportional to the information that can be extracted from dimension $i$ regarding category $j$. Based on this argument, Bhattacharya distance \cite{bhattacharyya1943} with normal distribution assumption is a suitable metric, which is given in \eqref{eq:bhattacharya}, to quantify the level of encoding in the word embedding dimensions. Normality of the embedding dimensions are tested using one-sample Kolmogorov-Smirnov test (KS test, Bonferroni corrected for multiple comparisons).

\begin{multline} \label{eq:bhattacharya}
{\mathcal{W}_B(i,j)} = \frac{1}{4}\ln\left(\frac{1}{4}\left(\frac{\sigma^2_{p_{i,j}}}{\sigma^2_{q_{i,j}}} + \frac{\sigma^2_{q_{i,j}}}{\sigma^2_{p_{i,j}}} + 2\right)\right) \\ +  \frac{1}{4}\left(\frac{\left(\mu_{p_{i,j}} - \mu_{q_{i,j}}\right)^2}{\sigma^2_{p_{i,j}} + \sigma^2_{q_{i,j}}}\right)
\end{multline}

In \eqref{eq:bhattacharya}, $\mathcal{W}_B$ is a $300\times110$ Bhattacharya distance matrix, which can also be considered as a category weight matrix, $i$ is the dimension index ($i \in \{1, 2, ..., 300\}$), $j$ is the category index ($j \in \{1, 2, ..., 110\}$). $p_{i,j}$ is the vector of the $i^{th}$ dimension of each word in $j^{th}$ category and $q_{i,j}$ is the vector of the $i^{th}$ dimension of all other vocabulary words ($p_{i,j}$ is of length $n_j$ and $q_{i,j}$ is of length ($50000 - n_j$) where $n_j$ is the number of words in the $j^{th}$ category). $\mu$ and $\sigma$ are the mean and the standard deviation operations, respectively. Values in $\mathcal{W}_B$ can range from $0$ (if $p_{i,j}$ and $q_{i,j}$ have the same means and variances) to $\infty$. In general, a better separation of category words from remaining vocabulary words in a dimension results in larger $\mathcal{W}_B$ elements for the corresponding dimension.

Based on SEMCAT categories, for the learned embedding matrices $\mathcal{E}$ and $\mathcal{E}^2$, the category weight matrices ($\mathcal{W}_B$ and $\mathcal{W}^2_B$) are calculated using Bhattacharya distance metric \eqref{eq:bhattacharya}. 

\subsection{Interpretable Word Vector Generation}\label{Interpretable Word Embeddings}

\begin{figure*}
	\centering
	\includegraphics[width=18.0cm]{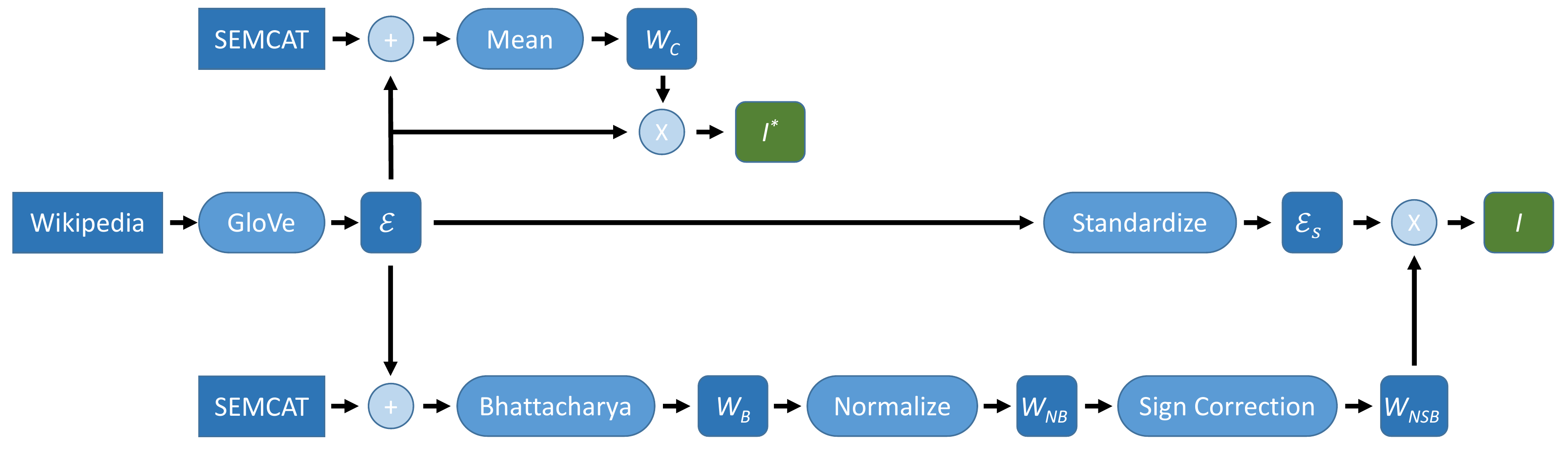}
	\caption{ Flow chart for the generation of the interpretable embedding spaces $\mathcal{I}$ and $\mathcal{I}^*$. First, word vectors are obtained using the GloVe algorithm on Wikipedia corpus. To obtain $\mathcal{I}^*$, weight matrix $\mathcal{W}_C$ is generated by calculating the means of the words from each category for each embedding dimension and then $\mathcal{W}_C$ is multiplied by the embedding matrix (see Section \ref{Interpretable Word Embeddings}). To obtain $\mathcal{I}$, weight matrix $\mathcal{W}_B$ is generated by calculating the Bhattacharya distance between category words and remaining vocabulary for each category and dimension. Then, $\mathcal{W}_B$ is normalized (see section \ref{Interpretable Word Embeddings} item 2), sign corrected (see section \ref{Interpretable Word Embeddings} item 3) and finally multiplied with standardized word embedding ($\mathcal{E}_s$, see Section \ref{Interpretable Word Embeddings} item 1)}
	\label{flow_chart}
\end{figure*} 

If the weights in $\mathcal{W}_B$ truly correspond to the categorical decomposition of the semantic concepts in the dense embedding space, then $\mathcal{W}_B$ can also be considered as a transformation matrix that can be used to map word embeddings to a semantic space where each dimension is a semantic category. However, it would be erroneous to directly multiply the word embeddings with category weights. The following steps should be performed in order to map word embeddings to a semantic space where dimensions are interpretable: 

\begin{enumerate}
	
	\item To make word embeddings compatible in scale with the category weights, word embedding dimensions are standardized ($\mathcal{E}_S$) such that each dimension has zero mean and unit variance since category weights have been calculated based on the deviations from the general mean (second term in \eqref{eq:bhattacharya}) and standard deviations (first term in \eqref{eq:bhattacharya}).
	
	\item Category weights are normalized across dimensions such that each category has a total weight of 1 ($\mathcal{W}_{NB}$). This is necessary since some columns of $\mathcal{W}_B$ dominate others in terms of representation strength (will be discussed in Section \ref{results} in more detail). This inequality across semantic categories can cause an undesired bias towards categories with larger total weights in the new vector space. $\ell_1$ normalization of the category weights across dimensions is performed to prevent bias. 
	
	\item Word embedding dimensions can encode semantic categories in both positive and negative directions ($\mu_{p_{i,j}} - \mu_{q_{i,j}}$ can be positive or negative) that contribute equally to the Bhattacharya distance. However, since encoding directions are important for the mapping of the word embeddings, $\mathcal{W}_{NB}$ is replaced with its signed version $\mathcal{W}_{NSB}$ (if $\mu_{p_{i,j}} - \mu_{q_{i,j}}$ is negative, then $\mathcal{W}_{NSB}(i,j) =  -\mathcal{W}_{NB}(i,j)$, otherwise $\mathcal{W}_{NSB}(i,j) = \mathcal{W}_{NB}(i,j)$) where negative weights correspond to encoding in the negative direction.
	
\end{enumerate}

Then, interpretable semantic vectors ($\mathcal{I}_{50000\times110}$) are obtained by multiplying $\mathcal{E}_S$ with $\mathcal{W}_{NSB}$. 

One can reasonably suggest to alternatively use the centers of the vectors of the category words as the weights for the corresponding category as given in (2).

\begin{equation} \label{Wc}
	\mathcal{W}_C(i,j)=\mu_{p_{i,j}}
\end{equation}

A second interpretable embedding space, $\mathcal{I}^*$, is then obtained by simply projecting the word vectors in $\mathcal{E}$ to the category centers. (3) and (4) show the calculation of $\mathcal{I}$ and $\mathcal{I}^*$ respectively. Figure \ref{flow_chart} shows the procedure for generation of interpretable embedding spaces $\mathcal{I}$ and $\mathcal{I}^*$.

\begin{eqnarray}
	\mathcal{I} = \mathcal{E}_S\mathcal{W}_{NSB} \\
	\mathcal{I}^* = \mathcal{E}\mathcal{W}_C
\end{eqnarray}

\subsection{Validation}\label{m:validation}

$\mathcal{I}$ and $\mathcal{I}^*$ are further investigated via qualitative and quantitative approaches in order to confirm that $\mathcal{W}_B$ is a reasonable semantic decomposition of the dense word embedding dimensions, that $\mathcal{I}$ is indeed an interpretable semantic space and that our proposed method produces better representations for the categories than their center vectors.

If $\mathcal{W}_B$ and $\mathcal{W}_C$ represent the semantic distribution of the word embedding dimensions, then columns of $\mathcal{I}$ and $\mathcal{I}^*$ should correspond to semantic categories. Therefore, each word vector in $\mathcal{I}$ and $\mathcal{I}^*$ should represent the semantic decomposition of the respective word in terms of the SEMCAT categories. To test this prediction, word vectors from the two semantic spaces ($\mathcal{I}$ and $\mathcal{I}^*$) are qualitatively investigated. 

To compare $\mathcal{I}$ and $\mathcal{I}^*$, we also define a quantitative test that aims to measure how well the category weights represent the corresponding categories. Since weights are calculated directly using word vectors, it is natural to expect that words should have high values in dimensions that correspond to the categories they belong to. However, using words that are included in the categories for investigating the performance of the calculated weights is similar to using training accuracy to evaluate model performance in machine learning. Using validation accuracy is more adequate to see how well the model generalizes to new, unseen data that, in our case, correspond to words that do not belong to any category. During validation, we randomly select 60\% of the words for training and use the remaining 40\% for testing for each category. From the training words we obtain the weight matrix $\mathcal{W}_B$ using Bhattacharya distance and the weight matrix $\mathcal{W}_C$ using the category centers. We select the largest $k$ weights ($k \in \{5,7,10,15,25,50,100,200,300\}$) for each category (i.e. largest $k$ elements for each column of $\mathcal{W}_B$ and $\mathcal{W}_C$) and replace the other weights with 0 that results in sparse category weight matrices ($\mathcal{W}_B^s$ and $\mathcal{W}_C^s$). Then projecting dense word vectors onto the sparse weights from $\mathcal{W}_B^s$ and $\mathcal{W}_C^s$, we obtain interpretable semantic spaces $\mathcal{I}_k$ and $\mathcal{I}^*_k$. Afterwards, for each category, we calculate the percentages of the unseen test words that are among the top $n$, $3n$ and $5n$ words (excluding the training words) in their corresponding dimensions in the new spaces, where $n$ is the number of test words that varies across categories. We calculate the final accuracy as the weighted average of the accuracies across the dimensions in the new spaces, where the weighting is proportional to the number of test words within the categories. We repeat the same procedure for 10 independent random selections of the training words.

\begin{figure}[t]
	\centering
	\includegraphics[width=9cm]{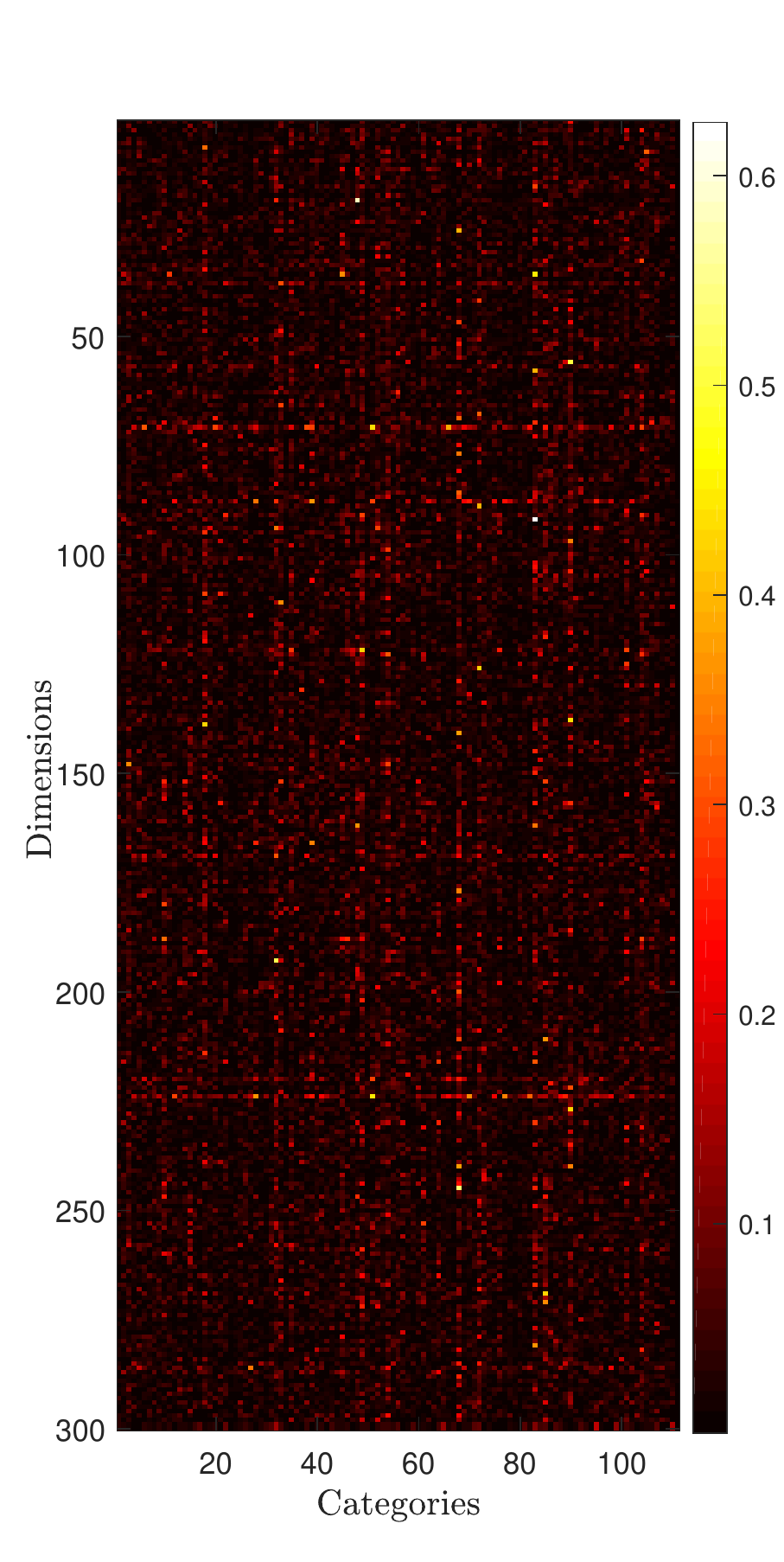}
	\caption{Semantic category weights ($\mathcal{W}_{B~300\times110}$) for 110 categories and 300 embedding dimensions obtained using Bhattacharya distance. Weights vary between 0 (represented by black) and 0.63 (represented by white). It can be noticed that some dimensions represent larger number of categories than others do and also some categories are represented strongly by more dimensions than others.}
	\label{cat weights}
\end{figure}

\begin{figure}[b]
	\centering
	\hspace*{-0.5cm}
	\includegraphics[width=10.0cm]{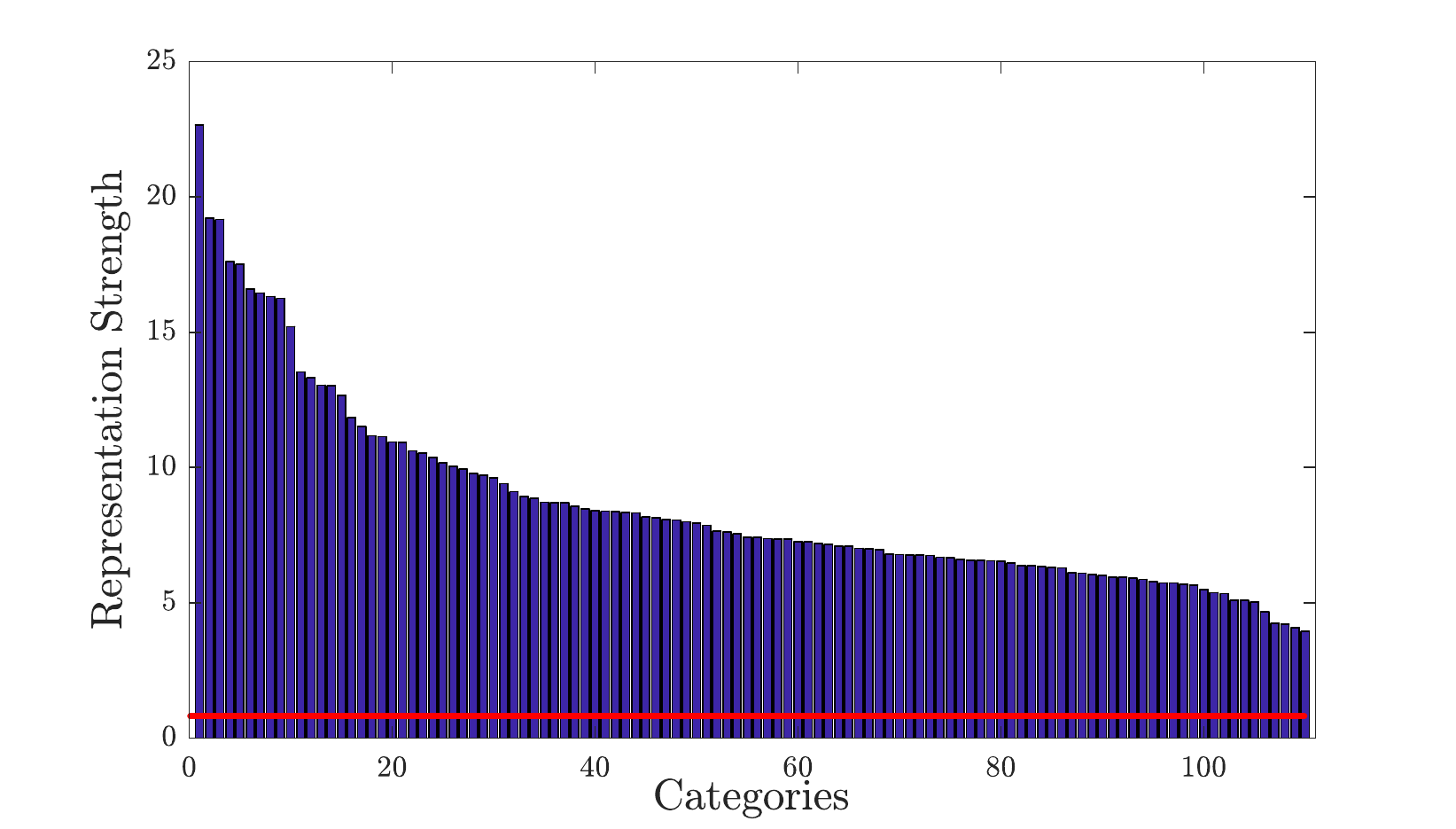}
	\caption{Total representation strengths of 110 semantic categories from SEMCAT. Bhattacharya distance scores are summed across dimensions and then sorted. Red horizontal line represents the baseline strength level obtained for a category composed of 91 randomly selected words from the vocabulary (where 91 is the average word count across categories in SEMCAT). The metals category has the strongest total representation among SEMCAT categories due to relatively few and well clustered words it contains while the pirate category has the lowest total representation due to widespread words it contains.}
	\label{total_reps}
\end{figure}

\begin{figure*}
	\centering
	\hspace*{-2cm}
	\includegraphics[width=8.9in]{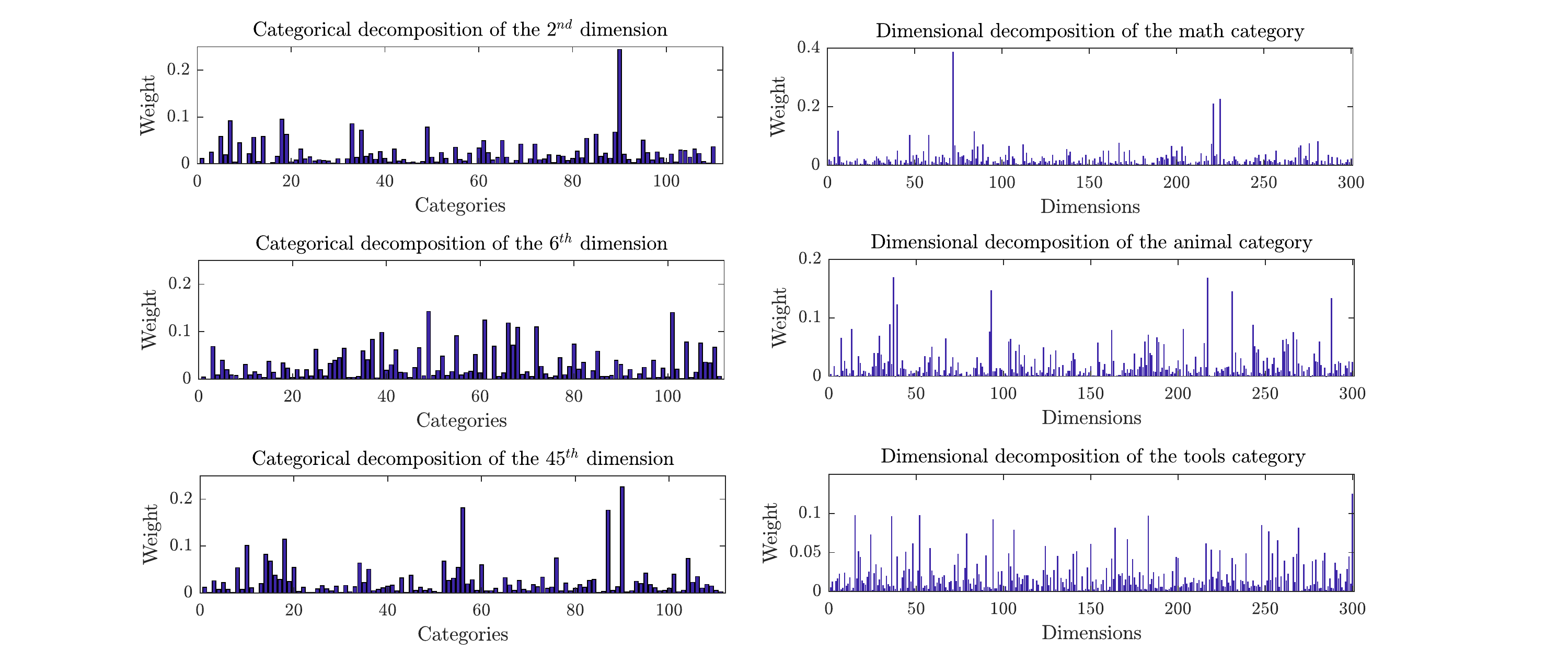}
	\caption{Categorical decompositions of the $2^{nd}$, $6^{th}$ and $45^{th}$ word embedding dimensions are given in the left column. A dense word embedding dimension may focus on a single category (top row), may represent a few different categories (bottom row) or may represent many different categories with low strength (middle row). Dimensional decompositions of the math, animal and tools categories are shown in the right column. Semantic information about a category may be encoded in a few word embedding dimensions (top row) or it can be distributed across many of the dimensions (bottom row).}
	\label{some_cat_weghts}
\end{figure*}

\subsection{Measuring Interpretability}

In addition to investigating the semantic distribution in the embedding space, a word category dataset can be also used to quantify the interpretability of the word embeddings. In several studies, \cite{murphy2012, luo2015, chang2009}, interpretability is evaluated using the word intrusion test. In the word intrusion test, for each embedding dimension, a word set is generated including the top 5 words in the top ranks and a noisy word (intruder) in the bottom ranks of that dimension. The intruder is selected such that it is in the top ranks of a separate dimension. Then, human editors are asked to determine the intruder word within the generated set. The editors' performances are used to quantify the interpretability of the embedding. Although evaluating interpretability based on human judgements is an effective approach, word intrusion is an expensive method since it requires human effort for each evaluation. Furthermore, the word intrusion test does not quantify the interpretability levels of the embedding dimensions, instead it yields a binary decision as to whether a dimension is interpretable or not. However, using continuous values is more adequate than making binary evaluations since interpretability levels may vary gradually across dimensions. 

We propose a framework that addresses both of these issues by providing automated, continuous valued evaluations of interpretability while keeping the basis of the evaluations as human judgements. The basic idea behind our framework is that humans interpret dimensions by trying to group the most distinctive words in the dimensions (i.e. top or bottom rank words), an idea also leveraged by the word intrusion test. Based on this key idea, it can be noted that if a dataset represents all the possible groups humans can form, instead of relying on human evaluations, one can simply check whether the distinctive words of the embedding dimensions are present together in any of these groups. As discussed earlier, the number of groups humans can form is theoretically unbounded, therefore it is not possible to compile an all-comprehensive dataset for all potential groups. However, we claim that a dataset with a sufficiently large number of categories can still provide a good approximation to human judgements. Based on this argument, we propose a simple method to quantify the interpretability of the embedding dimensions.

We define two interpretability scores for an embedding dimension-category pair as:

\begin{equation} \label{eq:interpretability_1}
\begin{split}
IS^+_{i,j}=\frac{|S_j \cap V^+_i(\lambda \times n_j)|}{n_j} \times 100  \\
IS^-_{i,j}=\frac{|S_j \cap V^-_i(\lambda \times n_j)|}{n_j} \times 100 
\end{split}
\end{equation} 

where $IS^+_{i,j}$ is the interpretability score for the positive direction and $IS^-_{i,j}$ is the interpretability score for the negative direction for the $i^{th}$ dimension ($i \in \{1,2,...,D\}$ where $D$ is the dimensionality of the embedding) and $j^{th}$ category ($j \in \{1,2,...,K\}$ where $K$ is the number of categories in the dataset). $S_j$ is the set representing the words in the $j^{th}$ category, $n_j$ is the number of the words in the $j^{th}$ category and $V^+_i(\lambda \times n_j)$, $V^-_i(\lambda \times n_j)$ refer to the distinctive words located at the top and bottom ranks of the $i^{th}$ embedding dimension, respectively. $\lambda \times n_j$ is the number of words taken from the upper and bottom ranks where $\lambda$ is the parameter determining how strict the interpretability definition is. The smallest value for $\lambda$  is 1 that corresponds to the most strict definition and larger $\lambda$ values relax the definition by increasing the range for selected category words. $\cap$ is the intersection operator between category words and top and bottom ranks words, $|.|$ is the cardinality operator (number of elements) for the intersecting set.

We take the maximum of scores in the positive and negative directions as the overall interpretability score for a category ($IS_{i,j}$). The interpretability score of a dimension is then taken as the maximum of individual category interpretability scores across that dimension ($IS_{i}$). Finally, we calculate the overall interpretability score of the embedding ($IS$) as the average of the dimension interpretability scores:

\begin{equation} \label{eq:interpretability_2}
\begin{split}
IS_{i,j} &= \max(IS^+_{i,j}, IS^-_{i,j}) \\
IS_{i} &= \max_{j} IS_{i,j}				\\
IS &= \frac{1}{D}\sum\limits_{i=1}^D IS_{i}
\end{split}
\end{equation}

We test our method on the GloVe embedding space, on the semantic spaces $\mathcal{I}$ and $\mathcal{I}^*$, and on a random space where word vectors are generated by randomly sampling from a zero mean, unit variance normal distribution. Interpretability scores for the random space are taken as our baseline. We measure the interpretability scores as $\lambda$ values are varied from 1 (strict interpretability) to 10 (relaxed interpretability).

Our interpretability measurements are based on our proposed dataset SEMCAT, which was designed to be a comprehensive dataset that contains a diverse set of word categories. Yet, it is possible that the precise interpretability scores that are measured here are biased by the dataset used. In general, two main properties of the dataset can affect the results: category selection and within-category word selection. To examine the effects of these properties on interpretability evaluations, we create alternative datasets by varying both category selection and word selection for SEMCAT. Since SEMCAT is comprehensive in terms of the words it contains for the categories, these datasets are created by subsampling the categories and words included in SEMCAT. Since random sampling of words within a category may perturb the capacity of the dataset in reflecting human judgement, we subsample r\%  of the words that are closest to category centers within each category, where $r \in \{40,60,80,100\}$. To examine the importance of number of categories in the dataset we randomly select $m$ categories from SEMCAT where $m \in \{30,50,70,90,110\}$. We repeat the selection 10 times independently for each $m$.  

\section{Results}\label{results}

\subsection{Semantic Decomposition}

The KS test for normality reveals that 255 dimensions of $\mathcal{E}$ are normally distributed ($p > 0.05$). The average test statistic for these 255 dimensions is $0.0064 \pm 0.0016$ (mean $\pm$ standard deviation). While the normality hypothesis was rejected for the remaining 45 dimensions, a relatively small test statistic of $0.0156 \pm 0.0168$ is measured, indicating that the distribution of these dimensions is approximately normal.

The semantic category weights calculated using the method introduced in Section \ref{m:semantic_decomposition} is displayed in Figure \ref{cat weights}. A close examination of the distribution of category weights indicates that the representation of semantic concepts are broadly distributed across many dimensions of the GloVe embedding space. This suggests that the raw space output by the GloVe algorithm has poor interpretability.

\begin{figure}
	\centering
	\hspace*{-0.5cm}
	\includegraphics[width=9.8cm]{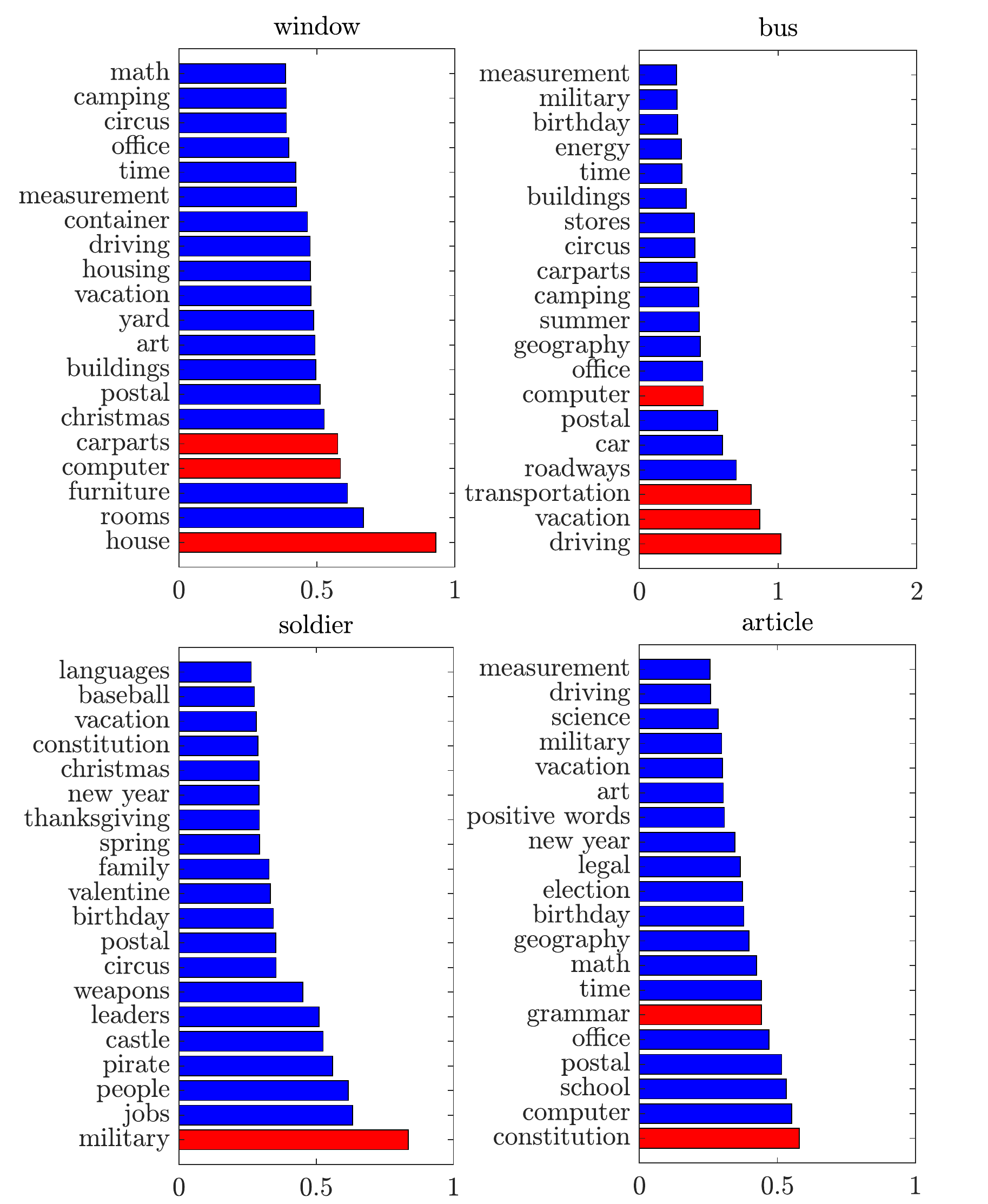}
	\caption{Semantic decompositions of the words \textit{window}, \textit{bus}, \textit{soldier} and \textit{article} for 20 highest scoring SEMCAT categories obtained from vectors in $\mathcal{I}$. Red bars indicate the categories that contain the word, blue bars indicate the categories that do not contain the word.}
	\label{fig:bhat_cat_decomps}
\end{figure}

\begin{figure}
	\centering
	\hspace*{-0.5cm}
	\includegraphics[width=9.8cm]{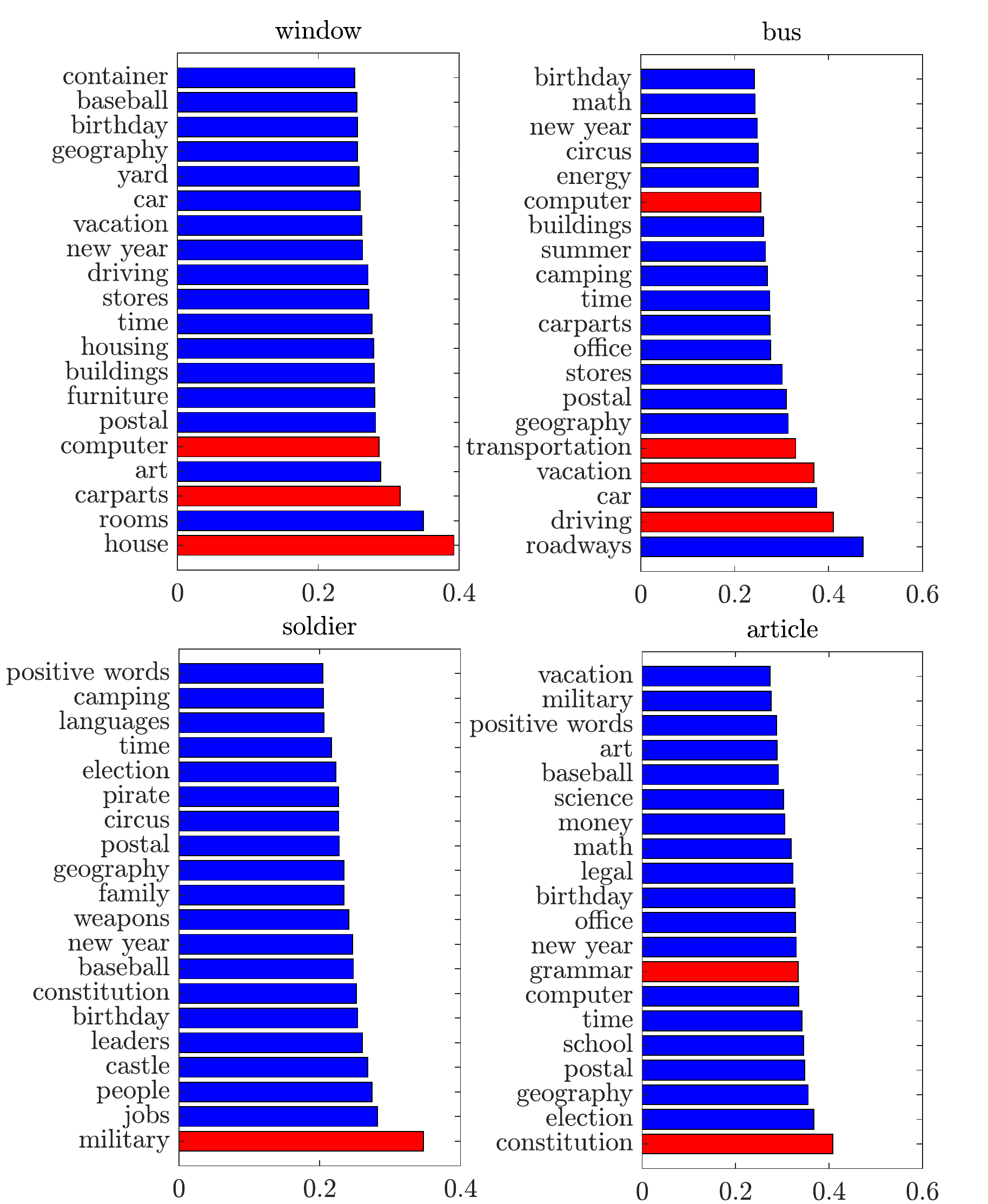}
	\caption{Categorical decompositions of the words \textit{window}, \textit{bus}, \textit{soldier} and \textit{article} for 20 highest scoring categories obtained from vectors in $\mathcal{I}^*$. Red bars indicate the categories that contain the word, blue bars indicate the categories that do not contain the word.}
	\label{fig:glove_cat_decomps}
\end{figure}

\begin{figure*}
	\centering
	\includegraphics[width=18cm]{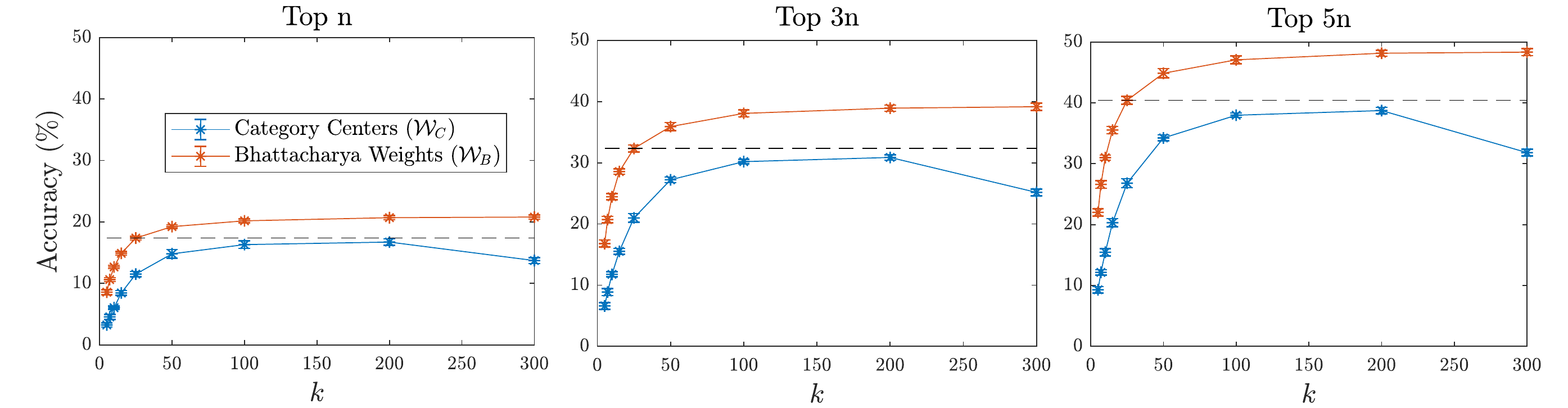}
	\caption{Category word retrieval performances for top $n$, $3n$ and $5n$ words where $n$ is the number of test words varying across categories. Category weights obtained using Bhattacharya distance represent categories better than the center of the category words. Using only 25 largest weights from $\mathcal{W}_B$ for each category ($k = 25$) gives better performance than using category centers with any $k$ (shown with dashed line).}
	\label{fig:cat_word_retrieval_test_results}
\end{figure*}

In addition, it can be observed that the total representation strength summed across dimensions varies significantly across categories, some columns in the category weight matrix contain much higher values than others. In fact, total representation strength of a category greatly depends on its word distribution. If a particular category reflects a highly specific semantic concept with relatively few words such as the metals category, category words tend to be well clustered in the embedding space. This tight grouping of category words results in large Bhattacharya distances in most dimensions indicating stronger representation of the category. On the other hand, if words from a semantic category are weakly related, it is more difficult for the word embedding to encode their relations. In this case, word vectors are relatively more widespread in the embedding space, and this leads to smaller Bhattacharya distances indicating that the semantic category does not have a strong representation across embedding dimensions. The total representation strengths of the 110 semantic categories in SEMCAT are shown in Figure \ref{total_reps}, along with the baseline strength level obtained for a category composed of 91 randomly selected words where 91 is the average word count across categories in SEMCAT. The metals category has the strongest total representation among SEMCAT categories due to relatively few and well clustered words it contains, whereas the pirate category has the lowest total representation due to widespread words it contains. 

To closely inspect the semantic structure of dimensions and categories, let us investigate the decompositions of three sample dimensions and three specific semantic categories (math, animal and tools). The left column of Figure \ref{some_cat_weghts} displays the categorical decomposition of the $2^{nd}$, $6^{th}$ and $45^{th}$ dimensions of the word embedding. While the $2^{nd}$ dimension selectively represents a particular category (sciences), the $45^{th}$ dimension focuses on 3 different categories (housing, rooms and sciences) and the $6^{th}$ dimension has a distributed and relatively uniform representation of many different categories. These distinct distributional properties can also be observed in terms of categories as shown in the right column of Figure \ref{some_cat_weghts}. While only few dimensions are dominant for representing the math category, semantic encodings of the tools and animals categories are distributed across many embedding dimensions.

Note that these results are valid regardless of the random initialization of the GloVe algorithm while learning the embedding space. For the weights calculated for our second GloVe embedding space $\mathcal{E}^2$, where the only difference between $\mathcal{E}$ and $\mathcal{E}^2$ is the independent random initializations of the word vectors before training, we observe nearly identical decompositions for the categories ignoring the order of the dimensions (similar number of peaks and similar total representation strength; not shown).

\subsection{Validation}\label{r:validation}

A representative investigation of the semantic space $\mathcal{I}$ is presented in Figure \ref{fig:bhat_cat_decomps}, where semantic decompositions of 4 different words, \textit{window}, \textit{bus}, \textit{soldier} and \textit{article}, are displayed using 20 dimensions of $\mathcal{I}$ with largest values for each word. These words are expected to have high values in the dimensions that encode the categories to which they belong. However, we can clearly see from Figure \ref{fig:bhat_cat_decomps} that additional categories such as jobs, people, pirate and weapons that are semantically related to \text{soldier} but that do not contain the word also have high values. Similar observations can be made for \textit{window}, \textit{bus}, and \textit{article} supporting the conclusion that the category weight spread broadly to many non-category words.

Figure \ref{fig:glove_cat_decomps} presents the semantic decompositions of the words \text{window}, \text{bus}, \text{soldier} and \text{article} obtained form $\mathcal{I}^*$ that is calculated using the category centers. Similar to the distributions obtained in $\mathcal{I}$, words have high values for semantically-related categories even when these categories do not contain the words. In contrast to $\mathcal{I}$, however, scores for words are much more uniformly distributed across categories, implying that this alternative approach is less discriminative for categories than the proposed method.

To quantitatively compare $\mathcal{I}$ and $\mathcal{I}^*$, category word retrieval test is applied and the results are presented in Figure \ref{fig:cat_word_retrieval_test_results}. As depicted in Figure \ref{fig:cat_word_retrieval_test_results}, the weights calculated using our method ($\mathcal{W}_B$) significantly outperform the weights from the category centers ($\mathcal{W}_C$). It can be noticed that, using only 25 largest weights from $\mathcal{W}_B$ for each category ($k = 25$) yields higher accuracy in word retrieval compared to the alternative $\mathcal{W}_C$  with any $k$. This result confirms the prediction that the vectors that we obtain for each category (i.e. columns of $\mathcal{W}_B$) distinguish categories better than their average vectors (i.e. columns of $\mathcal{W}_C$).

\subsection{Measuring Interpretability}\label{r:interpretability}

\begin{figure}[t]
	\centering
	\hspace*{-0.5cm}
	\includegraphics[width=10cm]{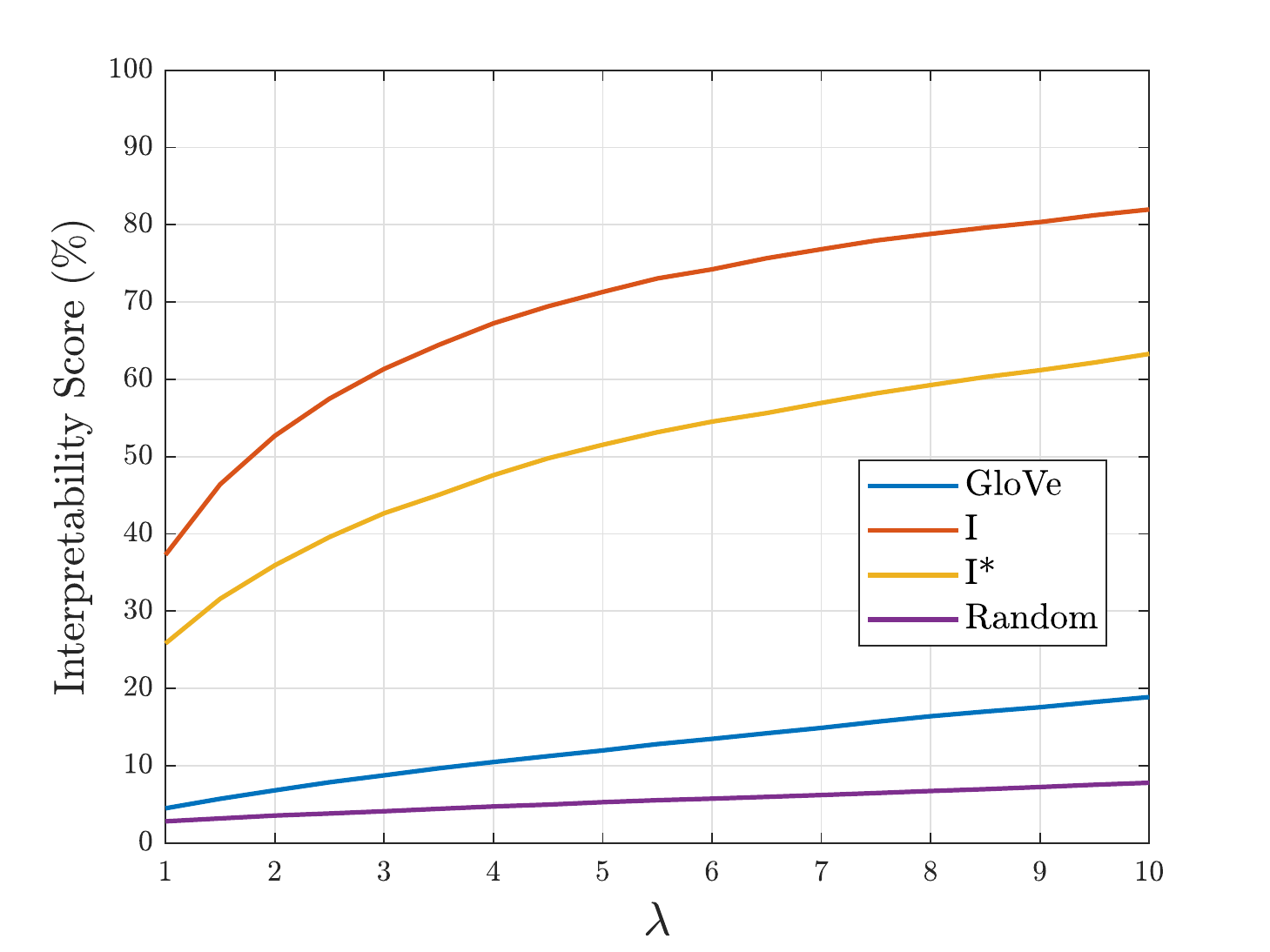}
	\caption{Interpretability scores for GloVe, $\mathcal{I}$, $\mathcal{I^*}$ and random embeddings for varying $\lambda$ values  where $\lambda$ is the parameter determining how strict the interpretability definition is ($\lambda = 1$ is the most strict definition, $\lambda = 10$ is a relaxed definition). Semantic spaces $\mathcal{I}$ and $\mathcal{I^*}$ are significantly more interpretable than GloVe as expected. $\mathcal{I}$ outperforms $\mathcal{I^*}$ suggesting that weights calculated with our proposed method more distinctively represent categories as opposed weights calculated as the category centers. Interpretability scores of Glove are close to the baseline (Random) implying that the dense word embedding has poor interpretability.}
	\label{interpretability}
\end{figure}

Figure \ref{interpretability} displays the interpretability scores of the GloVe embedding, $\mathcal{I}$, $\mathcal{I}^*$ and the random embedding for varying $\lambda$ values. $\lambda$ can be considered as a design parameter adjusted according to the interpretability definition. Increasing $\lambda$ relaxes the interpretability definition by allowing category words to be distributed on a wider range around the top ranks of a dimension. We propose that $\lambda = 5$ is an adequate choice that yields a similar evaluation to measuring the top-5 error in category word retrieval tests. As clearly depicted, semantic space $\mathcal{I}$ is significantly more interpretable than the GloVe embedding as justified in Section \ref{r:validation}. We can also see that interpretability score of the GloVe embedding is close to the random embedding representing the baseline interpretability level.

Interpretability scores for datasets constructed by sub-sampling SEMCAT are given in Table \ref{tab:interpetability} for the GloVe, $\mathcal{I}$, $\mathcal{I}^*$ and random embedding spaces for $\lambda = 5$. Interpretability scores for all embeddings increase as the number of categories in the dataset increase (30, 50, 70, 90, 110) for each category coverage (40\%, 60\%, 80\%, 100\%). This is expected since increasing the number of categories corresponds to taking into account human interpretations more substantially during evaluation. One can further argue that true interpretability scores of the embeddings (i.e. scores from an all-comprehensive dataset) should be even larger than those presented in Table \ref{tab:interpetability}. However, it can also be noticed that the increase in the interpretability scores of the GloVe and random embedding spaces gets smaller for larger number of categories. Thus, there is diminishing returns to increasing number of categories in terms of interpretability. Another important observation is that the interpretability scores of $\mathcal{I}$ and $\mathcal{I}^*$ are more sensitive to number of categories in the dataset than the GloVe or random embeddings. This can be attributed to the fact that $\mathcal{I}$ and $\mathcal{I}^*$ comprise dimensions that correspond to SEMCAT categories, and that inclusion or exclusion of these categories more directly affects interpretability.

In contrast to the category coverage, the effects of within-category word coverage on interpretability scores can be more complex. Starting with few words within each category, increasing the number of words is expected to more uniformly sample from the word distribution, more accurately reflect the semantic relations within each category and thereby enhance interpretability scores. However, having categories over-abundant in words might inevitably weaken semantic correlations among them, reducing the discriminability of the categories and interpretability of the embedding. Table \ref{tab:interpetability} shows that, interestingly, changing the category coverage has different effects on the interpretability scores of different types of embeddings. As category word coverage increases, interpretability scores for random embedding gradually decrease while they monotonically increase for the GloVe embedding. For semantic spaces $\mathcal{I}$ and $\mathcal{I}^*$, interpretability scores increase as the category coverage increases up to 80$\%$ of that of SEMCAT, then the scores decrease. This may be a result of having too comprehensive categories as argued earlier, implying that categories with coverage of around 80$\%$ of SEMCAT are better suited for measuring interpretability. However, it should be noted that the change in the interpretability scores for different word coverages might be effected by non-ideal subsampling of category words. Although our word sampling method, based on words' distances to category centers, is expected to generate categories that are represented better compared to random sampling of category words, category representations  might be suboptimal compared to human designed categories.

\begin{table}[t]
	\centering
	\caption{Average Interpretability Scores ($\%$) for $\lambda = 5$. Results are averaged across 10 independent selections of categories for each category coverage.}
	\label{tab:interpetability} 
	\begin{tabular}{c|c|l|c|c|c|c|c|} 
		
		\multicolumn{3}{c}{} & \multicolumn{5}{c}{Number of Categories}\\ 
		\cline{4-8} \multicolumn{3}{c|}{} & 30 & 50 & 70 & 90 & 110 \\ \cline{2-8}
		\multirow{16}{*}{\rot{Category Coverage (\%)}} 	
		& \multirow{4}{*}{40} 	& Random 			& 4.9 & 5.5 & 6.0 & 6.4 & 6.7 \\
		&						& GloVe	 			& 5.6 & 6.8 & 7.7 & 8.3 & 8.9 \\
		&						& $\mathcal{I}^*$	& 25.9 & 33.6 & 40.2 & 44.8 & 49.1 \\
		&						& $\mathcal{I}$		& 34.2 & 45.2 & 55.5 & 62.9 & 69.2 \\ \cline{2-8}
		& \multirow{4}{*}{60} 	& Random 			& 4.5 & 4.9 & 5.3 & 5.6 & 5.8 \\ 
		&						& GloVe	 			& 6.7 & 7.8 & 9.0 & 9.7 & 10.2 \\
		&						& $\mathcal{I}^*$	& 27.6 & 35.8 & 42.4 & 47.7 & 51.6 \\
		&						& $\mathcal{I}$		& 36.1 & 48.4 & 59.0 & 67.0 & 72.8 \\ \cline{2-8}
		& \multirow{4}{*}{80}	& Random 			& 4.2 & 4.6 & 4.9 & 5.1 & 5.3 \\
		&						& GloVe	 			& 7.6 & 8.9 & 9.7 & 10.4 & 11.0 \\
		&						& $\mathcal{I}^*$	& 30.2 & 31.1 & 43.2 & 48.1 & 52.0 \\
		&						& $\mathcal{I}$		& 39.8 & 50.7 & 60.1 & 67.4 & 73.2 \\ \cline{2-8}	
		& \multirow{4}{*}{100} 	& Random 			& 4.3 & 4.6 & 4.8 & 5.0 & 5.1 \\
		&						& GloVe	 			& 8.4 & 9.8 & 10.8 & 11.4 & 12.0 \\
		&						& $\mathcal{I}^*$	& 30.3 & 37.7 & 43.4 & 48.1 & 51.5 \\
		&						& $\mathcal{I}$		& 38.9 & 49.9 & 59.0 & 65.7 & 71.3 \\ \cline{2-8}		
	\end{tabular}
\end{table}


\section{Discussion and Conclusion}\label{conclusion}

In this paper, we propose a statistical method to uncover the latent semantic structure in dense word embeddings. Based on a new dataset (SEMCAT) we introduce that contains more than 6,500 words semantically grouped under 110 categories, we provide a semantic decomposition of the word embedding dimensions and verify our findings using qualitative and quantitative tests. We also introduce a method to quantify the interpretability of word embeddings based on SEMCAT that can replace the word intrusion test that relies heavily on human effort while keeping the basis of the interpretations as human judgements. 

Our proposed method to investigate the hidden semantic structure in the embedding space is based on calculation of category weights using a Bhattacharya distance metric. This metric implicitly assumes that the distribution of words within each embedding dimension is normal. Our statistical assessments indicate that the GloVe embedding space considered here closely follows this assumption. In applications where the embedding method yields distributions that significantly deviate from a normal distribution, nonparametric distribution metrics such as Spearman's correlation could be leveraged as an alternative. The resulting category weights can seamlessly be input to the remaining components of our framework. 

Since our proposed framework for measuring interpretability depends solely on the selection of the category words dataset, it can be used to directly compare different word embedding methods (e.g., GloVe, word2vec, fasttext) in terms of the interpretability of the resulting embedding spaces. A straightforward way to do this is to compare the category weights calculated for embedding dimensions across various embedding spaces. Note, however, that the Bhattacharya distance metric for measuring the category weights does not follow a linear scale and is unbounded. For instance, consider a pair of embeddings with category weights 10 and 30 versus another pair with weights 30 and 50. For both pairs, the latter embedding can be deemed more interpretable than the former. Yet, due to the gross nonlinearity of the distance metric, it is challenging to infer whether a 20-unit improvement in the category weights corresponds to similar levels of improvement in interpretability across the two pairs. To alleviate these issues, here we propose an improved method that assigns normalized interpretability scores with an upper bound of 100\%. This method facilitates interpretability assessments and comparisons among separate embedding spaces.

The results reported in this study for semantic analysis and interpretability assessment of embeddings are based on SEMCAT. SEMCAT contains 110 different semantic categories where average number of words per category is 91 rendering SEMCAT categories quite comprehensive. Although the HyperLex dataset contains a relatively larger number of categories (1399), the average number of words per category is only 2, insufficient to accurately represent semantic categories. Furthermore, while HyperLex categories are constructed based on a single type of relation among words (hyperonym-hyponym), SEMCAT is significantly more comprehensive since many categories include words that are grouped based on diverse types of relationships that go beyond hypernym-hyponym relations. Meanwhile, the relatively smaller number of categories in SEMCAT is not considered a strong limitation, as our analyses indicate that the interpretability levels exhibit diminishing returns when the number of categories in the dataset are increased and SEMCAT is readily yielding near optimal performance. That said, extended datasets with improved coverage and expert labeling by multiple observers would further improve the reliability of the proposed approach. To do this, a synergistic merge with existing lexical databases such as WordNet might prove useful.

Methods for learning dense word embeddings remain an active area of NLP research. The framework proposed in this study enables quantitative assessments on the intrinsic semantic structure and interpretability of word embeddings. Providing performance improvements in other common NLP tasks might be a future study. Therefore, the proposed framework can be a valuable tool in guiding future research on obtaining interpretable yet effective embedding spaces for many NLP tasks that critically rely on semantic information. For instance, performance evaluation of more interpretable word embeddings on higher level NLP tasks (i.e. sentiment analysis, named entity recognition, question answering) and the relation between interpretability and NLP performance can be worthwhile. 

\section*{Acknowledgment}
We thank the anonymous reviewers for their constructive and helpful comments that have significantly improved our paper.

This work was supported in part by a European Molecular Biology Organization Installation Grant (IG 3028), by a TUBA GEBIP fellowship, and by a BAGEP 2017 award of the Science Academy.

\bibliographystyle{IEEEtran}
\bibliography{NLP_bibliography}

\begin{thebibliography}{10}
\providecommand{\url}[1]{#1}
\csname url@samestyle\endcsname
\providecommand{\newblock}{\relax}
\providecommand{\bibinfo}[2]{#2}
\providecommand{\BIBentrySTDinterwordspacing}{\spaceskip=0pt\relax}
\providecommand{\BIBentryALTinterwordstretchfactor}{4}
\providecommand{\BIBentryALTinterwordspacing}{\spaceskip=\fontdimen2\font plus
\BIBentryALTinterwordstretchfactor\fontdimen3\font minus
  \fontdimen4\font\relax}
\providecommand{\BIBforeignlanguage}[2]{{%
\expandafter\ifx\csname l@#1\endcsname\relax
\typeout{** WARNING: IEEEtran.bst: No hyphenation pattern has been}%
\typeout{** loaded for the language `#1'. Using the pattern for}%
\typeout{** the default language instead.}%
\else
\language=\csname l@#1\endcsname
\fi
#2}}
\providecommand{\BIBdecl}{\relax}
\BIBdecl

\bibitem{miller1995wordnet}
G.~A. Miller, ``Wordnet: a lexical database for english,'' \emph{Communications
  of the ACM}, vol.~38, no.~11, pp. 39--41, 1995.

\bibitem{mikolov2013b}
T.~Mikolov, K.~Chen, G.~Corrado, and J.~Dean, ``Efficient estimation of word
  representations in vector space,'' \emph{arXiv preprint arXiv:1301.3781},
  2013.

\bibitem{bordes2012}
A.~Bordes, X.~Glorot, J.~Weston, and Y.~Bengio, ``Joint learning of words and
  meaning representations for open-text semantic parsing,'' in \emph{Artificial
  Intelligence and Statistics}, 2012, pp. 127--135.

\bibitem{harris1954}
Z.~S. Harris, ``Distributional structure,'' \emph{Word}, vol.~10, no. 2-3, pp.
  146--162, 1954.

\bibitem{firth1957}
J.~R. Firth, ``A synopsis of linguistic theory, 1930-1955,'' \emph{Studies in
  linguistic analysis}, 1957.

\bibitem{deerwester1990lsa}
S.~Deerwester, S.~T. Dumais, G.~W. Furnas, T.~K. Landauer, and R.~Harshman,
  ``Indexing by latent semantic analysis,'' \emph{J Am Soc Inf Sci}, vol.~41,
  no.~6, p. 391, 1990.

\bibitem{blei2003lda}
D.~M. Blei, A.~Y. Ng, and M.~I. Jordan, ``Latent dirichlet allocation,''
  \emph{J Mach Learn Res}, vol.~3, no. Jan, pp. 993--1022, 2003.

\bibitem{bojanowski2016fasttext}
P.~Bojanowski, E.~Grave, A.~Joulin, and T.~Mikolov, ``Enriching word vectors
  with subword information,'' \emph{arXiv preprint arXiv:1607.04606}, 2016.

\bibitem{pennington2014}
\BIBentryALTinterwordspacing
J.~Pennington, R.~Socher, and C.~D. Manning, ``Glove: Global vectors for word
  representation,'' in \emph{EMNLP}, 2014, pp. 1532--1543. [Online]. Available:
  \url{http://www.aclweb.org/anthology/D14-1162}
\BIBentrySTDinterwordspacing

\bibitem{lin2015POS}
C.-C. Lin, W.~Ammar, C.~Dyer, and L.~Levin, ``Unsupervised pos induction with
  word embeddings,'' \emph{arXiv preprint arXiv:1503.06760}, 2015.

\bibitem{sienvcnik2015NER}
S.~K. Sien{\v{c}}nik, ``Adapting word2vec to named entity recognition,'' in
  \emph{NODALIDA 2015, May 11-13, 2015, Vilnius, Lithuania}, no. 109.\hskip 1em
  plus 0.5em minus 0.4em\relax Link{\"o}ping University Electronic Press, 2015,
  pp. 239--243.

\bibitem{iacobacci2016Disambiguation}
I.~Iacobacci, M.~T. Pilehvar, and R.~Navigli, ``Embeddings for word sense
  disambiguation: An evaluation study.'' in \emph{ACL (1)}, 2016.

\bibitem{yu2017Sentiment}
L.-C. Yu, J.~Wang, K.~R. Lai, and X.~Zhang, ``Refining word embeddings for
  sentiment analysis,'' in \emph{EMNLP}, 2017, pp. 545--550.

\bibitem{senel2017crossLingual}
L.~K. \c{S}enel, V.~Y\"{u}cesoy, A.~Ko\c{c}, and T.~\c{C}ukur, ``Measuring
  cross-lingual semantic similarity across european languages,'' in \emph{TSP},
  2017.

\bibitem{levy2014dependency}
O.~Levy and Y.~Goldberg, ``Dependency-based word embeddings.'' in \emph{ACL
  (2)}, 2014, pp. 302--308.

\bibitem{lund1996HAL}
K.~Lund and C.~Burgess, ``Producing high-dimensional semantic spaces from
  lexical co-occurrence,'' \emph{Behavior Research Methods, Instruments, \&
  Computers}, vol.~28, no.~2, pp. 203--208, 1996.

\bibitem{goodman2016European}
B.~Goodman and S.~Flaxman, ``European union regulations on algorithmic
  decision-making and a" right to explanation",'' \emph{arXiv preprint
  arXiv:1606.08813}, 2016.

\bibitem{bruni2014multimodal}
E.~Bruni, N.-K. Tran, and M.~Baroni, ``Multimodal distributional semantics.''
  \emph{J. Artif. Intell. Res.(JAIR)}, vol.~49, no. 2014, pp. 1--47, 2014.

\bibitem{hill2016simlex}
F.~Hill, R.~Reichart, and A.~Korhonen, ``Simlex-999: Evaluating semantic models
  with (genuine) similarity estimation,'' \emph{Comput. Linguist.}, 2016.

\bibitem{murphy2004}
G.~Murphy, \emph{The big book of concepts}.\hskip 1em plus 0.5em minus
  0.4em\relax MIT press, 2004.

\bibitem{chang2009}
J.~Chang, S.~Gerrish, C.~Wang, J.~L. Boyd-Graber, and D.~M. Blei, ``Reading tea
  leaves: How humans interpret topic models,'' in \emph{NIPS}, 2009, pp.
  288--296.

\bibitem{murphy2012}
B.~Murphy, P.~Talukdar, and T.~Mitchell, ``Learning effective and interpretable
  semantic models using non-negative sparse embedding,'' in \emph{COLING},
  2012, pp. 1933--1950.

\bibitem{luo2015}
H.~Luo, Z.~Liu, H.-B. Luan, and M.~Sun, ``Online learning of interpretable word
  embeddings.'' in \emph{EMNLP}, 2015, pp. 1687--1692.

\bibitem{fyshe2014}
A.~Fyshe, P.~P. Talukdar, B.~Murphy, and T.~M. Mitchell, ``Interpretable
  semantic vectors from a joint model of brain-and text-based meaning,'' in
  \emph{ACL}, vol. 2014.\hskip 1em plus 0.5em minus 0.4em\relax NIH Public
  Access, 2014, p. 489.

\bibitem{arora2016}
S.~Arora, Y.~Li, Y.~Liang, T.~Ma, and A.~Risteski, ``Linear algebraic structure
  of word senses, with applications to polysemy,'' \emph{arXiv preprint
  arXiv:1601.03764}, 2016.

\bibitem{faruqui2015}
M.~Faruqui, Y.~Tsvetkov, D.~Yogatama, C.~Dyer, and N.~Smith, ``Sparse
  overcomplete word vector representations,'' \emph{arXiv preprint
  arXiv:1506.02004}, 2015.

\bibitem{zobnin2017rotations}
A.~Zobnin, ``Rotations and interpretability of word embeddings: the case of the
  russian language,'' \emph{arXiv preprint arXiv:1707.04662}, 2017.

\bibitem{park2017rotated}
S.~Park, J.~Bak, and A.~Oh, ``Rotated word vector representations and their
  interpretability,'' in \emph{EMNLP}, 2017, pp. 401--411.

\bibitem{jang2017}
K.-R. Jang and S.-H. Myaeng, ``Elucidating conceptual properties from word
  embeddings,'' \emph{SENSE 2017}, p.~91, 2017.

\bibitem{vulic2016hyperlex}
I.~Vuli{\'c}, D.~Gerz, D.~Kiela, F.~Hill, and A.~Korhonen, ``Hyperlex: A
  large-scale evaluation of graded lexical entailment,'' \emph{arXiv preprint
  arXiv:1608.02117}, 2016.

\bibitem{gladkova2016intrinsic}
A.~Gladkova, A.~Drozd, and C.~Center, ``Intrinsic evaluations of word
  embeddings: What can we do better?'' \emph{ACL 2016}, p.~36, 2016.

\bibitem{bhattacharyya1943}
A.~Bhattacharyya, ``On a measure of divergence between two statistical
  populations defined by their probability distribution,'' \emph{Bull. Calcutta
  Math. Soc}, 1943.

\end{thebibliography}

\vspace{5cm}

\end{document}